\documentclass{article}

\PassOptionsToPackage{numbers, compress}{natbib}

\usepackage[final]{nips_2016}

\usepackage[utf8]{inputenc} % allow utf-8 input
\usepackage[T1]{fontenc}    % use 8-bit T1 fonts
\usepackage{hyperref}       % hyperlinks
\usepackage{url}            % simple URL typesetting
\usepackage{booktabs}       % professional-quality tables
\usepackage{amsfonts}       % blackboard math symbols
\usepackage{nicefrac}       % compact symbols for 1/2, etc.
\usepackage{microtype}      % microtypography
\usepackage[usenames,dvipsnames]{color}

\usepackage{tikz}
\usetikzlibrary{datavisualization, arrows, positioning}

\title{Learning Operations on a Stack with Neural Turing Machines}

\author{
  Tristan Deleu\\
  Snips\\
  Paris, France \\
  \texttt{tristan.deleu@snips.ai}
  \And
  Joseph Dureau\\
  Snips\\
  Paris, France \\
  \texttt{joseph.dureau@snips.ai}
}

\begin{document}

\maketitle

\begin{abstract}
Multiple extensions of Recurrent Neural Networks (RNNs) have been proposed recently to address the difficulty of storing information over long time periods. In this paper, we experiment with the capacity of Neural Turing Machines (NTMs) to deal with these long-term dependencies on well-balanced strings of parentheses. We show that not only does the NTM emulate a stack with its heads and learn an algorithm to recognize such words, but it is also capable of strongly generalizing to much longer sequences.
\end{abstract}

\section{Introduction}
Although neural networks shine at finding meaningful representations of the data, they are still limited in their capacity to plan ahead, reason and store information over long time periods. Keeping track of nested parentheses in a language model, for example, is a particularly challenging problem for RNNs \cite{karpathy2015visualizing}. It requires the network to somehow memorize the number of unmatched open parentheses. In this paper, we analyze the ability of Neural Turing Machines (NTMs) to recognize well-balanced strings of parentheses. We show that even though the NTM architecture does not explicitely operate on a stack, it is able to emulate this data structure with its heads. Such a behaviour was unobserved on other simple algorithmic tasks \cite{DBLP:journals/corr/GravesWD14}.

After a brief recall of the Neural Turing Machine architecture in Section~\ref{sec:ntm}, we show in Section~\ref{sec:experiments} how the NTM is able to learn an algorithm to recognize strings of well-balanced parentheses, called \emph{Dyck words}. We also show how this model is capable to strongly generalize to longer sequences.

\section{Related Work}

\paragraph{Grammar induction}
Deep learning models are often trained on large datasets, generally extracted from real-world data at the cost of an expensive labeling step by some expert. In the context of Natural Language Processing, an alternative is to generate data from an artificial language, based on a predefined grammar. Historically, these \emph{formal languages} have been used to evaluate the theoretical foundations of RNNs \cite{siegelmann:1993}.

Hochreiter and Schmidhuber \cite{hochreiter1997long} tested their new Long Short-Term Memory (LSTM) on the \emph{embedded Reber language}, to show how their output gates can be beneficial. This behaviour was later extended to a variety of context-free and context-sensitive languages \cite{gers2001lstm}. However, as opposed to these previous works focused on character-level language modeling, here our task of interest is the \emph{membership problem}. This is a classification problem, where positive examples are generated by a given grammar, and negative examples are randomly generated with the same alphabet.

\paragraph{Differentiable memory}
To enhance their capacity to retain information, RNNs can be augmented with an explicit and differentiable memory module. Memory Networks and Dynamic Memory Networks \cite{DBLP:journals/corr/SukhbaatarSWF15,DBLP:journals/corr/MillerFDKBW16,DBLP:journals/corr/XiongMS16} use a hierarchical attention mechanism on an associative array to solve text QA tasks involving reasoning. 
Closely related our work, Stack-augmented RNNs \cite{DBLP:journals/corr/JoulinM15} are capable of inferring algorithmic patterns on some context-free and context-sensitive languages, including $a^{n}b^{n}$, $a^{n}b^{n}c^{n}$, and $a^{n}b^{m}c^{n+m}$.

\section{Neural Turing Machines}
\label{sec:ntm}

The Neural Turing Machine (NTM) \cite{DBLP:journals/corr/GravesWD14} is an instance of memory-augmented neural networks, consisting of a neural network \emph{controller} which interacts with a large (though bounded, unlike a Turing machine) \emph{memory tape}. The NTM uses soft read and write heads to retrieve information from the memory and store information in memory. The dynamics of these heads are governed by one or multiple sets of weights $\mathbf{w}^{r}_{t}$ for the read head(s) and $\mathbf{w}^{w}_{t}$ for the write head(s). These are controlled by the controller (either a Feed-forward network, or an LSTM), and maintain the overall architecture differentiable. The read head returns a read vector $\mathbf{r}_{t}$ as a weighted sum over the rows of the memory bank $M_{t}$:

\begin{equation}
     \mathbf{r}_{t} = \sum_{i}\mathbf{w}^{r}_{t}(i) M_{t}(i)
 \end{equation}

 Similarly, the write head modifies the memory $M_{t}$ by first erasing a weighted version of some \emph{erase vector} $\mathbf{e}_{t}$ from each row in the memory (Equation~\ref{eq:ntm:erase}), then adding a weighted version of an \emph{add vector} $\mathbf{a}_{t}$ (Equation~\ref{eq:ntm:add}). Both vectors $\mathbf{e}_{t}$ and $\mathbf{a}_{t}$ are generated by the controller.

 \begin{eqnarray}
     \tilde{M}_{t+1}(i) & \longleftarrow & M_{t}(i)\cdot(1 - \mathbf{w}^{w}_{t}(i)\,\mathbf{e}_{t}) \label{eq:ntm:erase}\\
     M_{t+1}(i) & \longleftarrow & \tilde{M}_{t+1}(i) + \mathbf{w}^{w}_{t}(i)\,\mathbf{a}_{t} \label{eq:ntm:add}
 \end{eqnarray}
 
The weights $\mathbf{w}^{r}_{t}$ and $\mathbf{w}^{r}_{t}$ are produced through a series of differentiable operations, called the addressing mechanisms. These fall into two categories: a \emph{content-based addressing} comparing each memory locations with some key $\mathbf{k}_{t}$, and a \emph{location-based addressing} responsible for shifting the heads (similar to a Turing machine). Even though recent works \cite{DBLP:journals/corr/SantoroBBWL16,DBLP:journals/corr/GulcehreCCB16} tend to drop the location-addressing, we chose to use the original formulation of the NTM and keep both addressing mechanisms.

\begin{figure}[ht]
    \centering
    \begin{tikzpicture}[xscale=4,yscale=3]
        \node[inner sep=0pt] (Ctrltm2) at (-0.75,0) {};
        \node[inner xsep=10pt, inner ysep=7pt, draw] (Ctrltm1) at (0,0) {Controller};
        \node[inner xsep=10pt, inner ysep=7pt, draw] (Ctrlt) at (1,0) {Controller};
        \begin{scope}[node distance=3pt]
            \node[circle, fill=black,inner sep=1pt] (n3) [left=of Ctrlt] {\tiny \color{white} \textbf{3}};
        \end{scope}
        \node[inner xsep=10pt, inner ysep=7pt, draw] (Ctrltp1) at (2,0) {Controller};
        \node[inner sep=0pt] (Mtm2) at (-0.75,-0.5) {};
        \node[inner sep=7pt] (Mtm1) at (-0.5,-0.5) {$M_{t-1}$};
        \node[inner sep=7pt] (Mt) at (0.5,-0.5) {$M_{t}$};
        \node[inner sep=7pt] (Mtp1) at (1.5,-0.5) {$M_{t+1}$};
        \node[inner sep=7pt] (Mtp2) at (2.5,-0.5) {$M_{t+2}$};
        \node[inner sep=0pt] (Mtp3) at (2.75,-0.5) {};
        \node[inner sep=7pt] (xtm1) at (0,-1) {$\mathbf{x}_{t-1}$};
        \node[inner sep=7pt] (xt) at (1,-1) {$\mathbf{x}_{t}$};
        \node[inner sep=7pt] (xtp1) at (2,-1) {$\mathbf{x}_{t+1}$};
        \node[inner sep=7pt] (htm1) at (0,0.35) {};
        \node (ht) at (1,0.35) {};
        \node (htp1) at (2,0.35) {};
        \node[inner sep=0pt,label=below:\tiny$\mathbf{r}_{t-1}$] (rtm1) at (-0.25,-0.75) {};
        \node[inner sep=0pt,label=below:\tiny$\mathbf{r}_{t}$] (rt) at (0.75,-0.75) {};
        \node[inner sep=0pt,label=below:\tiny$\mathbf{r}_{t+1}$] (rtp1) at (1.75,-0.75) {};
        \node[inner sep=0pt] (rtp2) at (2.75,-0.75) {};
        \draw[densely dashed,-stealth] (Mtm2) -- (Mtm1);
        \draw[densely dashed,-stealth] (Mtm1) -- (Mt);
        \draw[densely dashed,-stealth] (Mt) -- (Mtp1);
        \draw[densely dashed,-stealth] (Mtp1) -- (Mtp2);
        \draw[densely dashed] (Mtp2) -- (Mtp3);
        \draw[-stealth, rounded corners=3pt,thick] (Ctrltm2.east) -| (Mtm1.north);
        \draw[-stealth, rounded corners=3pt,thick] (Ctrltm1.east) -| (Mt.north);
        \draw[-stealth, rounded corners=3pt,thick] (Ctrlt.east) -| (Mtp1.north) node [midway, above right=2pt, circle, fill=black,inner sep=1pt] {\tiny \color{white} \textbf{4}};
        \draw[-stealth, rounded corners=3pt,thick] (Ctrltp1.east) -| (Mtp2.north);
        \draw[-stealth,thick] (Ctrltm1) -- (htm1);
        \draw[-stealth,thick] (Ctrlt) -- (ht);
        \draw[-stealth,thick] (Ctrltp1) -- (htp1);
        \draw[-stealth,thick] (xtm1) -- (Ctrltm1);
        \draw[-stealth,thick] (xt) -- (Ctrlt);
        \draw[-stealth,thick] (xtp1) -- (Ctrltp1);
        \draw[rounded corners=3pt,thick] (Mtm1.south) |- (rtm1.north) -| (Ctrltm1.south);
        \draw[rounded corners=3pt,thick] (Mt.south) |- (rt.north) node [midway, below left=2pt, circle, fill=black,inner sep=1pt] {\tiny \color{white} \textbf{1}} -| (Ctrlt.south) node [midway, right=4pt, circle, fill=black,inner sep=1pt] {\tiny \color{white} \textbf{2}};
        \draw[rounded corners=3pt,thick] (Mtp1.south) |- (rtp1.north) -| (Ctrltp1.south);
        \draw[rounded corners=3pt,thick] (Mtp2.south) |- (rtp2.north);
    \end{tikzpicture}
    \caption{\emph{A Neural Turing Machine, unrolled in time} -- (1) The read head first returns a read vector $\mathbf{r}_{t}$ which is (2) concatenated with the input $\mathbf{x}_{t}$. Both vectors are sent to (3) the controller (either a Feed-forward network, or an LSTM) which is responsible for the computation of the internal state of the NTM, as well as the read and write heads. (4) This write head is then used to makes changes to the memory $M_{t+1}$.}
\end{figure}
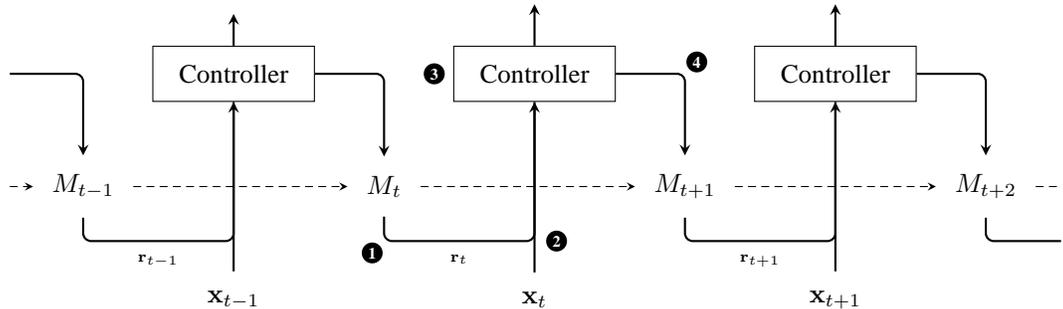

\section{Experiments}
\label{sec:experiments}

\subsection{Dyck words}
\label{sec:dyck}
\begin{figure}[ht]
\centering
\begin{tikzpicture}[anchor=base,xscale=0.70,yscale=0.70]
  \draw [help lines, densely dotted] (0,1) grid (14,4);
  \def\dyckexample{1,1,-1,1,1,-1,-1,-1,1,-1,1,1,-1,-1}
  \draw (0,1) \foreach \x in \dyckexample {-- ++(1,\x)};
  \draw (0,1) -- (14,1);
  \fill (0,1) circle (1.5pt);
  \draw (0,1) -- ++(0,1.5pt);
  \foreach \x [
    count=\i,
    remember=\height as \lastheight (initially 0),
    evaluate=\x as \height using \lastheight+\x,
    evaluate=\x as \ud using {\ifnum \x=1 {"u"} \else {"d"} \fi},
    evaluate=\x as \parenthesis using {\ifnum \x=1 {"("} \else {")"} \fi}
  ] in \dyckexample {
    \fill (\i,\height+1) circle (1.5pt);
    \node at (\i-0.5,0.5) {\small\ud};
    \node at (\i-0.5,0) {\small\parenthesis};
    \draw (\i,1) -- ++(0,1.5pt);
  }
\end{tikzpicture}
\caption{\emph{Example of a Dyck word} -- This is an example of a well-balanced string of parentheses (bottom), along with its representation in $\mathcal{A}^{*}$ (middle) and graphical representation as a path (top).}
% Example in D_14
\label{fig:dyck:dyck-path}
\end{figure}
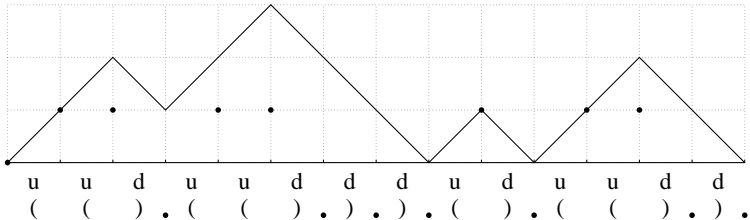

A \emph{Dyck word} is a balanced string of opening and closing parentheses. Besides the important role they play in parsing, they have multiple connections with other combinatorial objects \cite{stanley:1999,flajolet:2009}. In particular, one convenient and visual way representation of a Dyck word is a path on the integer line (see Figure~\ref{fig:dyck:dyck-path}).

To avoid ambiguities, we will consider strings of parentheses as words $w \in \left\{u, d\right\}^{*} = \mathcal{A}^{*}$, where each character $u$ corresponds to an opening parenthesis and $d$ to a closing parenthesis. The subset of $\mathcal{A}^{*}$ containing the Dyck words of length $<2n$ is called the \emph{Dyck language} and is denoted $\mathcal{D}_{<2n}$.

\subsection{Experimental setup}
\label{sec:setup}
We are interested here in the membership problem over the Dyck language. We trained a NTM for a classification task, where positive examples are uniformly sampled \cite{flajolet:2009} from the Dyck language $\mathcal{D}_{<12}$, and negative examples are non-Dyck words $w\in\mathcal{A}^{*}$ of length $<12$ with the same number of characters $u$ and $d$. We use the same experimental setup as described in \cite{DBLP:journals/corr/GravesWD14}, with a 1-layer feed-forward controller with 100 hidden units, 1 read head, 1 write head, and a memory bank containing 128 memory locations, each of dimension 20. We used a ReLU nonlinearity for the key $\mathbf{k}_{t}$ and add vector $\mathbf{a}_{t}$ and a hard sigmoid for the erase vector $\mathbf{e}_{t}$. We trained the model using the Adam optimizer \cite{DBLP:journals/corr/KingmaB14} with a learning rate of $0.001$ and batch size $16$.

\subsection{Stack emulation}
\begin{figure}[ht]
    \centering
    \includegraphics[width=0.90\textwidth]{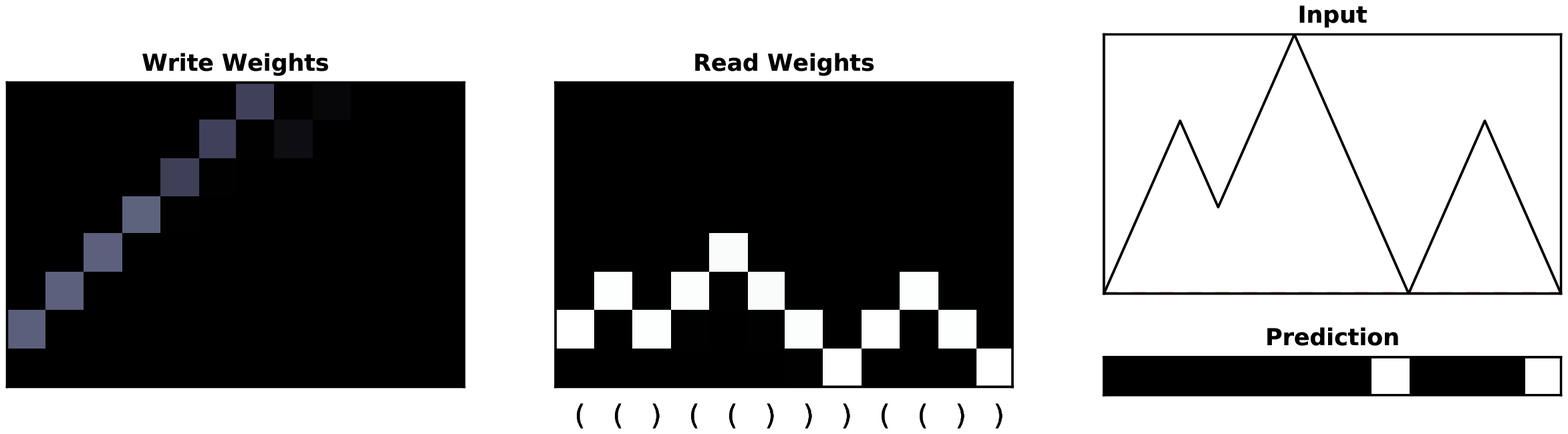}
    \includegraphics[width=0.90\textwidth]{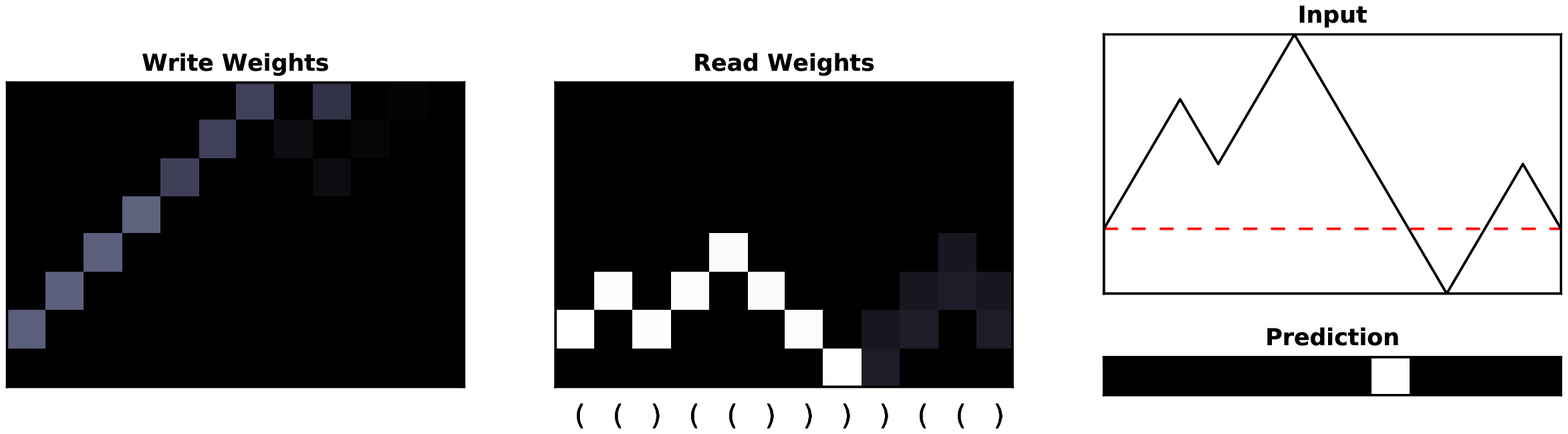}
    \caption{\emph{Read and write heads of the NTM} -- Examples of the behaviour of the NTM on a Dyck word (top) and a non-Dyck word (bottom) of length 12. For each example, we show the write (left) and read (center) weights as the NTM reads the input string.}
    \label{fig:experiments:ntm-dyck}
\end{figure}

The Dyck language is a context-free language that can be recognized by a pushdown automaton \cite{autebert:1997}. Here, we are interested in the nature of the algorithm the NTM is able to infer only from examples on this task. More specifically, we want to know if, and how, the NTM uses its memory to act as a stack, without specifying the push and pop operations explicitely \cite{DBLP:journals/corr/JoulinM15,DBLP:journals/corr/GrefenstetteHSB15}. In Figure~\ref{fig:experiments:ntm-dyck}, we show the behaviour of the read and write heads on a Dyck word and a non-Dyck word, along with the probability returned by the model of each prefix to be a Dyck word.

We observe that the model is actually emulating a stack with its read head. Each time the NTM reads an opening parenthesis $u$, the read head is moved upward and conversely when reading a closing parenthesis $d$. This behaviour is different from what was previously reported on other algorithmic tasks \cite{DBLP:journals/corr/GravesWD14}, where the content of the memory played a central role. Here, the NTM barely writes anything in memory, but uses its read head for computation purposes, following closely the graphical representation of the words (on the right).

The NTM uses its read head similarly for non-Dyck words, up until it reads a closing parenthesis with no matching opening parenthesis (illustrated by the red line in the graphical representation of the word), where the model correctly predicts the word is no longer a Dyck word. Beyond simply counting opening and closing parentheses, the NTM was also able to remember that violation point, despite the lack of recurrent controller.

\subsection{Strong generalization}
When testing a model, it is often assumed that the training and test data are sampled from the same (unknown) distribution. However, here we are not only interested in the capacity of the NTM to recognize Dyck words of similar length, but also its capacity to learn an algorithm and generalize to longer sequences. This is called \emph{strong generalization} \cite{DBLP:journals/corr/ReedF15}.

\begin{figure}[ht]
    \centering
    \includegraphics[width=0.7\textwidth]{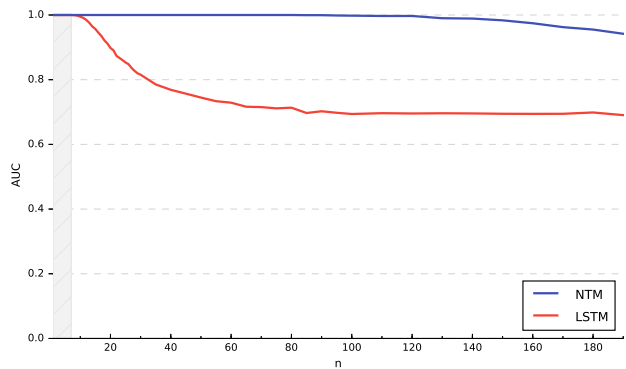}
    \caption{\emph{Generalization on $\mathcal{D}_{<2n}$} -- Strong generalization performance of a NTM (blue) and an LSTM with 10 hidden units (red) on sequences in $\mathcal{D}_{<2n}$, for different values of $n$. The gray area represents the training regime ($\mathcal{D}_{<12}$) for both models. The performance is reported as the Area Under the Curve (AUC).}
    \label{fig:experiments:auc}
\end{figure}

In Figure~\ref{fig:experiments:auc}, we compare the generalization performance of the NTM against an LSTM. This LSTM was selected as the model yielding the best AUC on sequences in $\mathcal{D}_{<200}$, with the number of hidden units selected in $[2, 5, 10, 20, 50, 100, 200, 500]$ (best: $10$). While the LSTM shows signs of strong generalization on sequences twice as long as what it was given during training, the AUC starts dropping for much longer sequences. On the other hand, the NTM generalizes perfectly even for much longer sequences (up to 20 times longer than the training regime). Beyond $n\approx 120$, the AUC starts to slightly decrease, most likely due to overflow issues: the stack emulated by the read head is limited by the number of memory locations, here 128.

\section{Conclusion}
Through an experiment on an artificial language called the Dyck language, we have shown that Neural Turing Machines are not only able to use their memory for storage, but can also use their heads for computational purposes. This allows the NTM to strongly generalize to inputs much longer, effectively learning an algorithm (contrary to only learning patterns in the data). The size of the memory allocated for the NTM being the only constraint. An interesting line of research could then be to run a similar experiment on a model trained under a memory-restricted regime, like a single memory location, and see how the NTM can emulate a stack under this stronger constraint.

\clearpage
\bibliography{dyckwords}
\bibliographystyle{plain}

\end{document}